\documentclass[journal]{IEEEtran}

\usepackage{cite}

\ifCLASSINFOpdf
  \usepackage[pdftex]{graphicx}
  \graphicspath{{../images/}}
  \DeclareGraphicsExtensions{.pdf,.jpeg,.png}
\else

\fi
\usepackage{amsmath}
\usepackage{algorithmic}
\usepackage{array}

\ifCLASSOPTIONcompsoc
 \usepackage[caption=false,font=normalsize,labelfont=sf,textfont=sf]{subfig}
\else
 \usepackage[caption=false,font=footnotesize]{subfig}
\fi

\usepackage{url}
\usepackage{color}
\usepackage{multirow}
\usepackage{booktabs}
\usepackage{adjustbox}
\usepackage{array}
\usepackage[inline,shortlabels]{enumitem}

\usepackage[switch]{lineno}  %

\begin{document}

\title{Stuttering Speech Disfluency Prediction \\using Explainable Attribution Vectors \\of Facial Muscle Movements}


\author{Arun~Das,
        Henry~Chacon, and
        Peyman~Najafirad, Secure AI and Autonomy Laboratory\\
		Jeffrey~Mock,
        Farzan~Irani, and
        Edward~Golob, Cognitive Neuroscience Laboratory
        
\thanks{This work has been submitted to the IEEE for possible publication. Copyright may be transferred without notice, after which this version may no longer be accessible. This work was partly supported by National Institutes of Health (NIH) under Grant DC016353 and the Open Cloud Institute (OCI) at University of Texas at San Antonio (UTSA). The authors gratefully acknowledge the use of the services of Jetstream cloud.}
\thanks{A. Das, H. Chacon, and P. Najafirad are members of the Secure AI and Autonomy Laboratory, University of Texas at San Antonio, San Antonio, TX, 78249 USA. e-mail: (arun.das, henry.chacon, peyman.najafirad)@utsa.edu}
\thanks{P. Najafirad is also with the Department of Information Systems and Cyber Security, University of Texas at San Antonio, San Antonio, TX, 78249 USA.}%
\thanks{J. Mock and E. Golob are with the Department of Psychology, University of Texas at San Antonio, San Antonio, TX, 78249 USA. e-mail: (jeffrey.mock, edward.golob)@utsa.edu.}%
\thanks{F. Irani is with the Department of Communication Disorders, Texas State University, San Marcos, TX, 78665 USA. e-mail: firani@txstate.edu}%
\thanks{J. Mock, E. Golob, and F. Irani are also members of the Cognitive Neuroscience Laboratory, University of Texas at San Antonio, San Antonio, TX, 78249 USA.}
}%

\markboth{IEEE TRANSACTIONS ON NEURAL SYSTEMS AND REHABILITATION ENGINEERING,~Vol.~X, No.~X, MONTH~YEAR}%
{Das \MakeLowercase{\textit{et al.}}: Stuttering Speech Disfluency Prediction using Explainable Attribution Vectors of Facial Muscle Movements}


\maketitle

\begin{abstract}
Speech disorders such as stuttering disrupt the normal fluency of speech by involuntary repetitions, prolongations and blocking of sounds and syllables. In addition to these disruptions to speech fluency, most adults who stutter (AWS) also experience numerous observable secondary behaviors before, during, and after a stuttering moment, often involving the facial muscles. Recent studies  have  explored  automatic detection of stuttering using Artificial Intelligence (AI) based algorithm from respiratory rate, audio, etc. during speech utterance. However, most methods require controlled environments and/or invasive wearable sensors, and are unable explain why a decision (fluent vs stuttered) was made. We hypothesize that pre-speech facial activity in AWS, which can be captured non-invasively, contains enough information to accurately classify the upcoming utterance as either fluent or stuttered. Towards this end, this paper proposes a novel explainable AI (XAI) assisted convolutional neural network (CNN) classifier to predict near future stuttering by learning temporal facial muscle movement patterns of AWS and explains the important facial muscles and actions involved. Statistical analyses reveal significantly high prevalence of cheek muscles (p$<$0.005) and lip muscles (p$<$0.005) to predict stuttering and  shows a behavior conducive of arousal  and  anticipation  to  speak.  The temporal study of these upper and lower facial  muscles may facilitate early detection of stuttering, promote automated assessment of stuttering and have application in behavioral therapies by providing automatic non-invasive feedback in realtime.
\end{abstract}

\begin{IEEEkeywords}
stuttering, speech disfluency, deep learning, explainable artificial intelligence, machine learning
\end{IEEEkeywords}

%
\IEEEpeerreviewmaketitle

\section{Introduction}
\label{sec:intro}
\IEEEPARstart{D}{evelopmental} stuttering is speech fluency disorder that begins in early childhood, with a reported incidence of 5\% and prevalence of 1\% in adults. Approximately 80\% of children recovery within 15 months from onset. For those children who persist and continue to stutter into adulthood, stuttering is characterized by the presence of excessive disfluencies interrupting the forward flow of speech. Specifically stuttering-like disfluencies (SLDs) include silent blocks, sound/syllable prolongations and part-word repetitions \cite{manning2017clinical}. SLDs are rarely observed in the speech of those who do not stutter. For adults who stutter (AWS), these SLDs (or disfluency) are often accompanied by numerous secondary behaviors that can be observed, before, during, and after the production of a SLD \cite{prasse2008stuttering}. Advances in our understanding of the neurophysiological underpinnings of stuttering have established  that stuttering is associated with structural and functional differences in the speech motor planning and programming regions of the brain. Previous research in our lab using EEG has demonstrated that neural oscillations occurring before the speaker attempts to speak can be used to predict whether an AWS will speak fluently  or stutter with 81\% classification accuracy \cite{Myers2019}. Using the same speech preparation paradigm, the central hypothesis of our current line of work is that at a given moment in time, the brain of an AWS is in a probabilistic state that exists between the extremes of certain generation of fluent or stuttered speech i.e.  p(fluent speech) $\in (0,1)$. We hypothesize that facial muscle activity can be used as an external marker to index internal brain states (see Myers et al., 2018 \cite{Myers2019}). We will test this hypothesis by determining if facial muscle activity measures, processed by convolutional neural networks (CNNs), can accurately predict if upcoming speech in AWS will be fluent or disfluent. We hypothesize that facial muscle activity can index fluent/disfluent brain states in AWS for the following three reasons.

First, pioneering work by Conture et al. [5] successfully distinguished non- speech facial motor activity patterns in children who stutter vs. fluent controls, even though only imprecise electromyogram (EMG) measures were available. Additionally, providing EMG feedback of facial muscle activity patterns increases the effectiveness of speech therapy in school-age children who stutter [6]. 
Second, there are plausible neuroanatomical mechanisms for why facial muscles may show different activation patterns before fluent/disfluent speech. Facial areas of the motor cortex body map in the brain are located right next to areas controlling speech articulators. Classic ideas suggest “motor overflow” is a general phenomenon in neurological disorders, whereby imprecise motor control at the level of motor cortex generates actions beyond those that were intended [7]. Relatedly, AWS also have deficits in motor sequence learning [8]. Third, emotion and affect are important aspects of stuttering [9], and facial microexpressions have a rich [10], although controversial history of indexing  emotions [11].

The activity of specific facial muscles can be quantified as Action Units (AUs), which are based on the Facial Action Unit Coding System (FACS) of Ekman and colleagues \cite{ekman1997face,friesen1978facial}. Facial AUs have been used extensively to study affect and spontaneous expressions \cite{Sayette2002,Hamm2011,Lints-Martindale2007}. Numerous studies have expressed the correlations and dynamics between emotional expression and upper facial muscle movements \cite{Wang2013,Wehrle2000,Mehu2012}. Also, lower AUs around the lips have been shown to be precise enough to classify the specific phoneme sequence created while speaking \cite{meng2017listen}. This study tries to take the next step and use facial muscle activity during speech preparation to distinguish whether an upcoming vocalization will be fluent or stuttered in AWS. Up until now, all automated analysis for distinguishing fluent vs. stuttered speech has been done with the aim to classify words within a recorded speech sample as fluent or stuttered by using either auditory vocalization data \cite{Zhang2013}, or respiratory rates \cite{Villegas2019}. To the best of our knowledge, we present the first work that uses pre-speech facial muscle movement to predict future behavior (fluent or stuttered speech, within approximately 2 to 3 seconds). The proposed method presents a non-invasive way of identifying pre-speech facial muscle movements that can classify future speech as either fluent or stuttered in AWS. 

We hypothesize that (1) pre-speech facial activity in AWS contain enough information to accurately classify near future speech as either fluent or stuttered, (2) while AWS prepare to speak, AUs can independently distinguish levels of speech-motor preparation and emotion related to speech output, (3) upper facial AUs could better index arousal and anticipation, and show greater activity in stuttered vs. fluent speech trials, (4) lower facial AUs will better index the level of motor preparation and lower facial AUs will show peak activity the closer one gets to speech onset, will be greater for stuttered vs. fluent speech trials, and show greater activity when a specific speech plan is held in memory. We test our hypothesis with rigorous experimental and statistical analyses.


Our major contributions are towards evaluating these hypotheses and can be summarized as below:
\begin{itemize}
\item We developed an experimental methodology to collect facial muscles before speech vocalization in AWS.
\item We  propose the use of a novel DL architecture based    on Convolution Neural Network (CNN) to classify future speech as either fluent or stuttered using pre-speech facial muscle movement activity from FACS.
\item To explain the classifier decisions, we will use and integrate DeepSHAP explainable AI algorithm to generate temporal explanation maps describing the significance of each AU over time.
\end{itemize}
The remaining manuscript is organized as follows: Section II provides summaries of related works and a background review. Section III describes the experimental design for human subject study, the AI, and XAI methods. Section IV, presents the experimental results of the proposed methods and statistical analysis. In Section V, we discuss the implications and impact of the results. Finally, in Section VI, we conclude the study and share future directions.

\section{Background Review}
\label{sec:relatedwork}

\subsection{Existing AI-based Stuttering Classifiers}
Myers et al. showed that neural oscillations occurring before the speaker attempts to speak can be used to predict whether an AWS will speak fluently or stutter with 81\% classification accuracy \cite{Myers2019}. Audio data \cite{Teixeira2017} and respiratory biosignals \cite{Villegas2019} have been used for classifying speech utterances that contain a silent motor block in individuals who stutter. In \cite{Villegas2019}, authors trained a multi-layer perceptron to differentiate block and non-block states from respiration rate and got 82.6\% classification accuracy on a test set omitted from training. However, the area-under-curve, precision, and recall values for the test was not included. The study also used invasive data collection methods due to requirement of elastic bands around the chest for respiratory inductance plethysmography (RIP) method and pulse oximeter.

An audio based automated stuttering classifier is described by Teixeira et al. in \cite{Teixeira2017}. The authors showed that the percentage of silence to speech in audio recordings could separate fluent from disfluent speech in people who stutter by using several methods such as zero crossing rate, moving average, and moving energy. However, authors recorded that the degree of speech disfluency is much harder to classify only with a measure of silence and speech. Work done by Mahesha et al. \cite{Mahesha2017} suggested better classifier performance using Mel Frequency Cepstral Coefficients (MFCC) to represent speech audio data. Audio based stuttering classifiers, though non-invasive, require a quiet environment for optimum performance as described in \cite{Teixeira2017}. This fundamental limitation discourages its use in a real-world environment with ambient noise. Further, these methods only detect stuttering at the time it happens and have not been tested in real-time. Our current study takes a large step forward from the above studies by using non-invasive facial AUs to determine if pre-speech facial movements can predict whether the upcoming utterance in AWS will be fluent or stuttered.

\subsection{Explaining AI Classifier Decisions}
Recent research has shed light on the importance of explainability in AI \cite{das2020opportunities}. The highly non-linear nature of deep learning algorithms restricts its use in mission critical applications such as healthcare. There is a need to explain the responsible features for a particular AI decision \cite{holzinger2017we}. In \cite{lundberg2017unified}, authors described SHAP, a game-theoretic method to explain feature importance towards particular AI decisions. Recent research in medical domain showed positive results in the use of SHAP to supplement model decisions \cite{tan20193d}. DeepSHAP method, a variant of SHAP, generates an explanation map, also called an attribution map, which describes the relevance of each input feature towards the AI decision. Thus, by studying the explanation map, we could evaluate the influence of AUs over time that correspond to near future speech disfluencies in AWS.

\section{Methods}
\label{sec:methods}
This section describes in detail the experimental design for separating speech preparation from speech production in AWS, the AI, and XAI frameworks. To aid the readers, we split this section to discuss the experimental design and protocols, DL classifier details, and design considerations for DeepSHAP explainability.

\subsection{Experimental Design and Protocol}
The current experimental design was based off our prior research that investigated how EEG data could identify moments during speech preparation that both differ between AWS and fluent adults \cite{mock2015speech,jeffmock2016cortical}, and can be used to classify brain states conducive to fluent or stuttered speech in AWS \cite{Myers2019}. We target speech preparation since over 90\% of disfluent events occur on the initial sound/syllable of an utterance \cite{sheehan1974stuttering}. In this study, we focus on non-invasive data collection methods using cameras to understand the role of facial muscle activity during speech preparation to classify stuttering disfluency. A certified speech-language pathologist was consulted to categorize individual trials to either fluent or stuttered.

\subsubsection{Subjects}
Seven male AWS (age=23.3$\pm$4 years, range 18-31 years) participated in the study. Each subject self-reported stuttering onset occurred during childhood and a certified speech-language pathologist diagnosed all of them with persistent developmental stuttering. On separate days, subjects completed between 1-5 sessions each (avg=3.7$\pm$1.7 sessions). Subjects have different numbers of sessions because we didn't start collecting video of the experimental stimuli (i.e. camera behind the subject) until midway through the data collection of 12 subjects (11 completing 5 sessions and 1 completing 3 sessions). Each participant signed a consent form, and studies were done in accordance to the protocols approved by The University of Texas at San Antonio Institutional Review Board, consistent with the Declaration of Helsinki.

\subsubsection{Hardware-Software Design}
\begin{figure}[!t]
\centering
\includegraphics[width=\columnwidth]{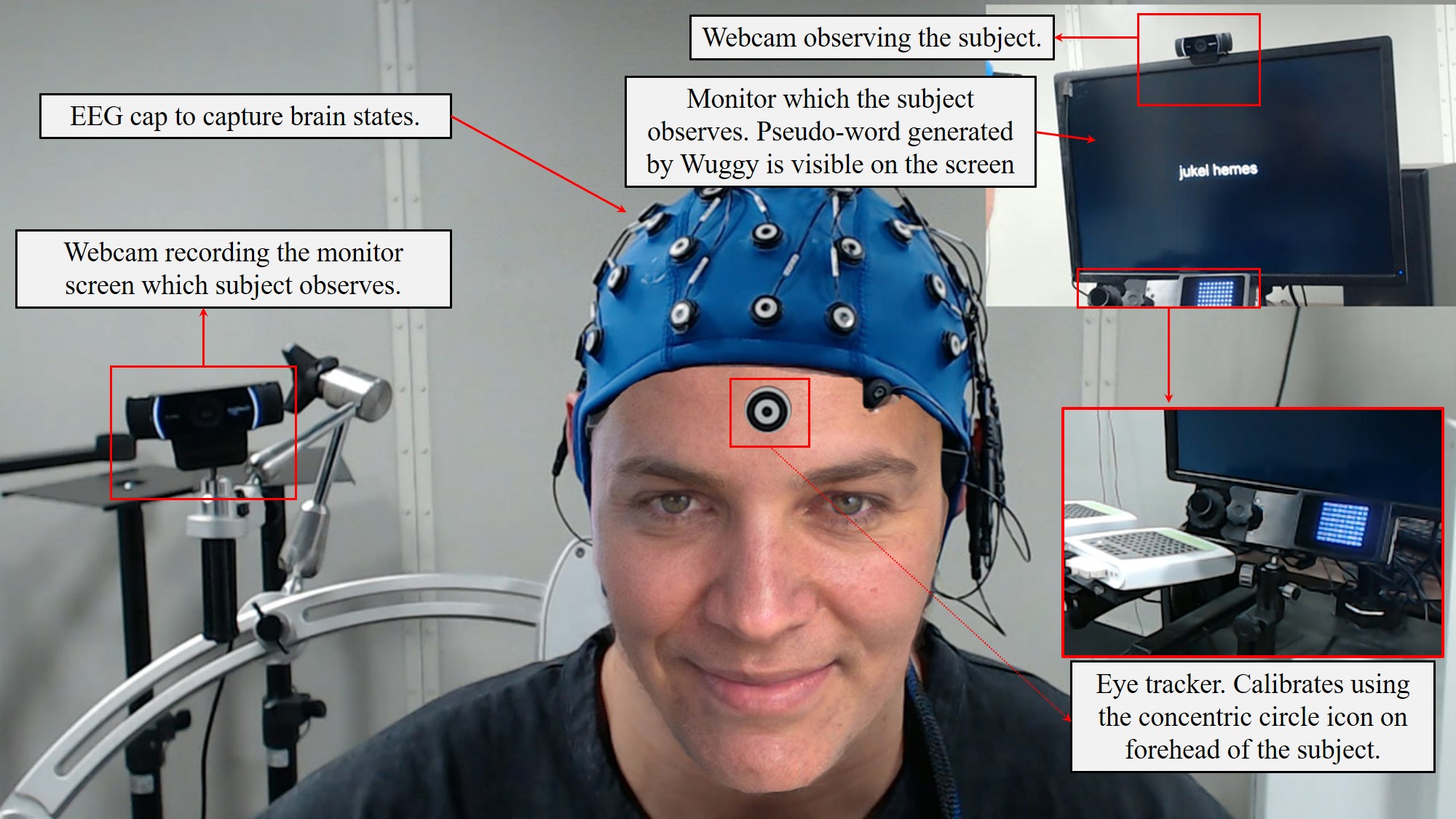}
\caption{A subject undergoing the S1-S2 task is illustated. Facial muscle activities are collected using a camera and the monitor which the subject views is also captured using another web-camera. Various other sensors such as EEG, eye-tracker, etc. collect other physiological and behavioral signals of the subject. The face of the actual subject is replaced by a face generated using an AI algorithm for visualization purposes available from thispersondoesnotexist.com.}
\label{fig:subjectdatacollection}
\end{figure}

\begin{figure}[!b]
\centering
\includegraphics[width=\columnwidth]{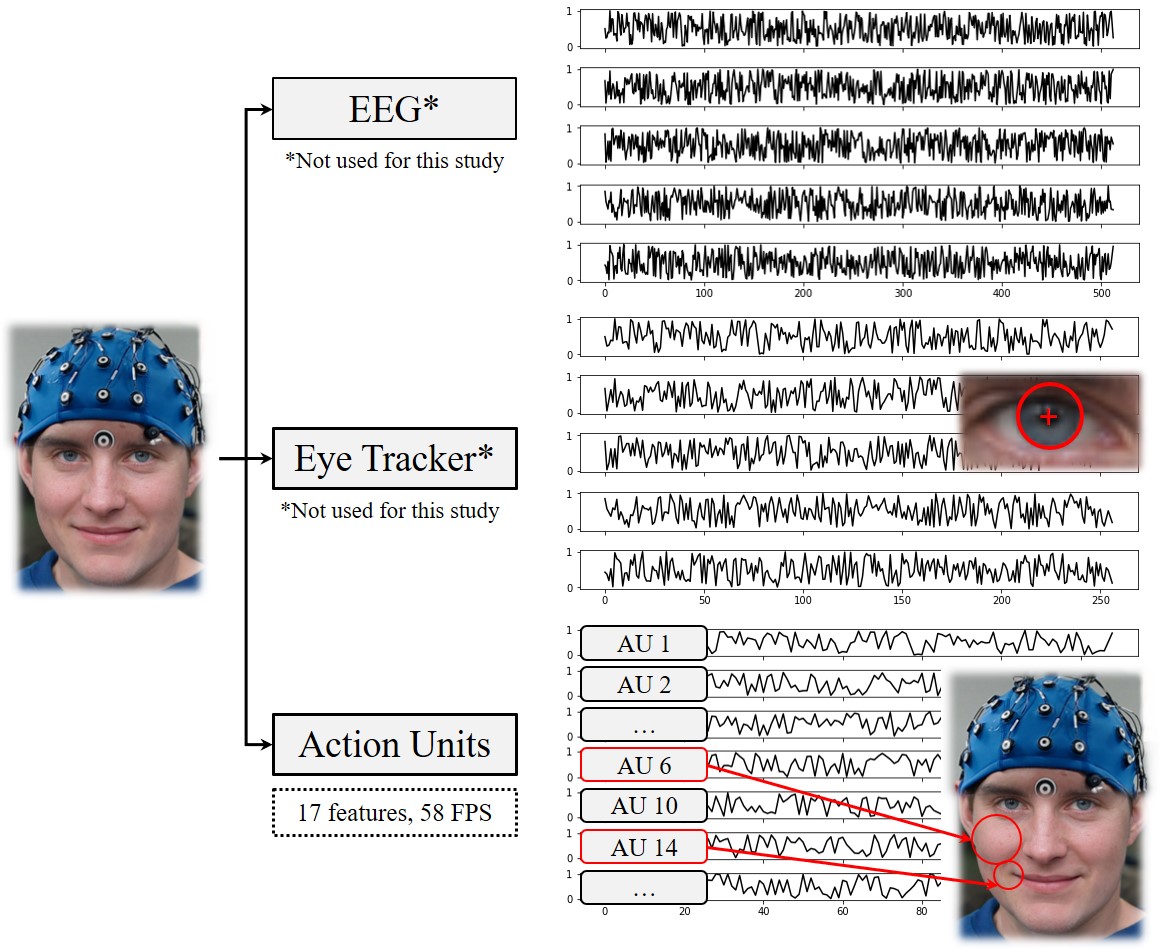}
\caption{Preprocessed sample data collected from the platform is illustrated. EEG, ET, and AU data are collected from each subject during each study session. However, for this research, we focus on the AU data. The face of the actual subject is replaced by a face generated using an AI algorithm for visualization purposes available from thispersondoesnotexist.com.}
\label{fig:sampledatacollected}
\end{figure}

To capture facial muscle movements in relation to the onset of the experimental stimuli, the face of the subject and the monitor used to present the visual stimuli were both recorded with two high-resolution cameras (Logitech C920 HD Pro) at 58 frames-per-second (FPS). The software ManyCam was used to combine both video streams into one continuous recording. Each recording was stored on encrypted UTSA assets according to subject, study, paradigm, and block information. Both 64 channels of EEG (Compumedics Neuroscan, Charlotte, NC) and high-resolution eye-tracking data (EyeLink 1000 Plus, SR research) were also recorded during all sessions. See Figure \ref{fig:subjectdatacollection} for a picture of the experimental setup and Figure \ref{fig:sampledatacollected} for various signals collected during the experiment. The current publication focuses solely on the facial AUs obtained by video.  

\subsubsection{Task and Paradigms}
\begin{figure*}[!t]
\centering
\includegraphics[width=\linewidth]{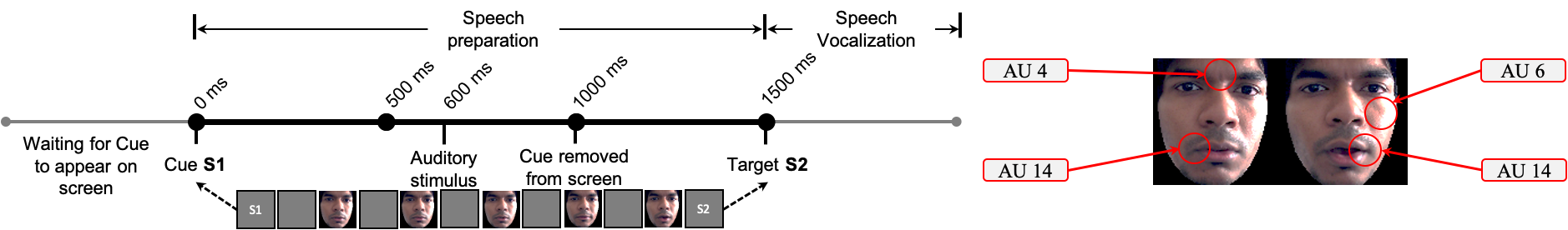}
\caption{Fundamental paradigms and their timing diagram is illustrated. Here, S1 represents the Cue where the trial begins. S2 represents the Target where a vocal response from subject is requested. CW and WG represent Cue-Word and Word-Go paradigms. CAW and WAG represent Cue-Auditory-Word and Word-Auditory-Go with additional 1000 Hz pure tone acoustic stimulus at 600 ms. Changes in the facial muscles of first author is illustrated as an example.}
\label{fig:paradigms}
\end{figure*}

To separate speech preparation from production, we imposed a delay between preparation and production by using  a `S1-S2' task. Our S1-S2 task provided a 1500 ms speech preparation period between S1 and S2 with the subject speaking after S2 onset as illustrated in Figure \ref{fig:paradigms}. Data was collected using four variations of the ‘S1-S2’ task with 100 trials of each variation (50 trials/block, 2 blocks/task, 400 trial/session).

For the duration of the experiment, each subject was seated in a sound booth that consisted of a computer monitor in front of them and 2 speakers directly to the left and right (24" away from head midline). Each trial started with a S1 visual stimulus (1000 ms duration), followed by 500 ms of a blank screen before the onset of a S2 visual stimulus (1000 ms duration) that signaled the subject to speak. In two of the task variations, named Word-Go (WG) and Word-Auditory-Go (WAG), S1 was a non-word pair and the S2 was three explanation points (!!!). The WG/WAG variations provided maximum information between S1-S2 on what was to be spoken after S2 onset. In the other two variations, Cue-Word (CW) and Cue-Auditory-Word (CAW), S1 consisted of a cross (+) while S2 was a non-word pair. The CW/CAW variations provided no information between S1-S2 on what was to be spoken except on the timing of S2 onset that told the subject to speak. The WAG/CAW task variations presented a tone (1000 Hz, 200 ms duration) 600 ms after S1, a time range our previous study found to correlate between EEG responses and individual differences in stuttering rate \cite{jeffmock2016cortical}. Due to no evidence that facial AUs are influenced by the auditory probe, the WAG and CAW tasks were grouped with the WG and CW tasks, respectively for this study. 

All tasks used non-word pairs (e.g. `jukel hemes'). Non-words were used because they phonetically mimic English words, have no associated meaning or emotional connotation, and allow for a more equal ratio of fluent to stuttered trials by increasing stuttering frequency above the typical 10\% average \cite{bloodstein2008handbook}. At the beginning of each session, a MATLAB script randomly selected 400 non-word pairs from a list of 900 possible non-words (avg. 2.1$\pm$0.6 syllables) with the limitation that the first and second non-words don't start with the same letter. This step produced unique non-word pairs on each trial and session. 
Offline, each trial was coded by a certified speech-language pathologist that specializes in developmental stuttering (4th author). Each trial was coded from the continuous video/ audio recordings as either fluent, unambiguous stuttering, ambiguous/normal disfluency (i.e. interjection such as `um'), or missed. A stuttered trial (unambiguous stuttering) was defined as either a silent block, a sound prolongation, or a part-word repetition. Only trials coded as fluent or stuttered were used in data analysis. 

\subsection{Face Muscle Movement Feature Extraction}

\subsubsection*{Face AU Extraction}
In order to extract facial action units, we used the model described in \cite{Baltrusaitis2015} with dynamic regressors which generates 17 facial action units from an incoming stream of video. The AI model to detect AUs is trained on different datasets including BP4D, SEMAINE, and DISFA, and generates person specific normalized action units with a generalized support vector regression (SVR) model. Hence, for each input frame, we get a vector of 17 elements representing 17 different action units from different regions of the face. We split the AU regions to upper face and lower facial regions. Specifically, AUs 1, 2, 4, 5, 6, 7, 9, and 45 constitute the upper facial region while AUs 10, 12, 14, 15, 17, 20, 23, 25, and 26 constitute the lower facial region.

\subsubsection*{Preprocessing}
At 58 FPS, the pre-speech information for one word utterance, as illustrated in Figure \ref{fig:paradigms}, accounts for 1.5 seconds of information between S1 and S2. This translates to 87 temporal data points. Considering 17 AUs extracted, we preprocess individual trials as a $17\times87$ matrix normalized in the range of 0 and 1. 

\subsubsection*{Dataset Generation for Deep Learning}

Individual subject data can be considered independent of each other. Also, since each trial involves a generated non-word, individual trials can be also considered independent of each other despite similarities between some AUs. A similar design choice was also done in \cite{samad2017feasibility}. Now, the goal of the CNN algorithm is to learn temporal dynamics between individual fluent and stuttered trials. After collecting face AU data from all subjects as matrices, we concatenate all AUs to generate the final datasets for deep learning study. Then, we split the AU datasets to training, test, and validation sets. 
Final dataset statistics is tabulated in Table \ref{tab:data_stat}.

\begin{table}[!b]
\caption{Dataset statistics of Action Unit (AU) data is tabulated. PAR, FEAT, and TS indicate paradigm, number of features, and timesteps, respectively.}
\label{tab:data_stat}
\centering
\begin{tabular}{|lllllll|}
\hline
PAR   & FEAT       & TS          & Total         & Train   & Test   & Validation \\
\hline
\hline
All   & 17         & 87          & 3704          & 2370    & 741    & 593          \\
CW    & 17         & 87          & 1710          & 1094    & 342    & 135          \\
WG    & 17         & 87          & 1992          & 1214    & 339    & 319          \\
\hline
\end{tabular}
\end{table}

\begin{figure*}[!t]
\centering
\includegraphics[width=\linewidth]{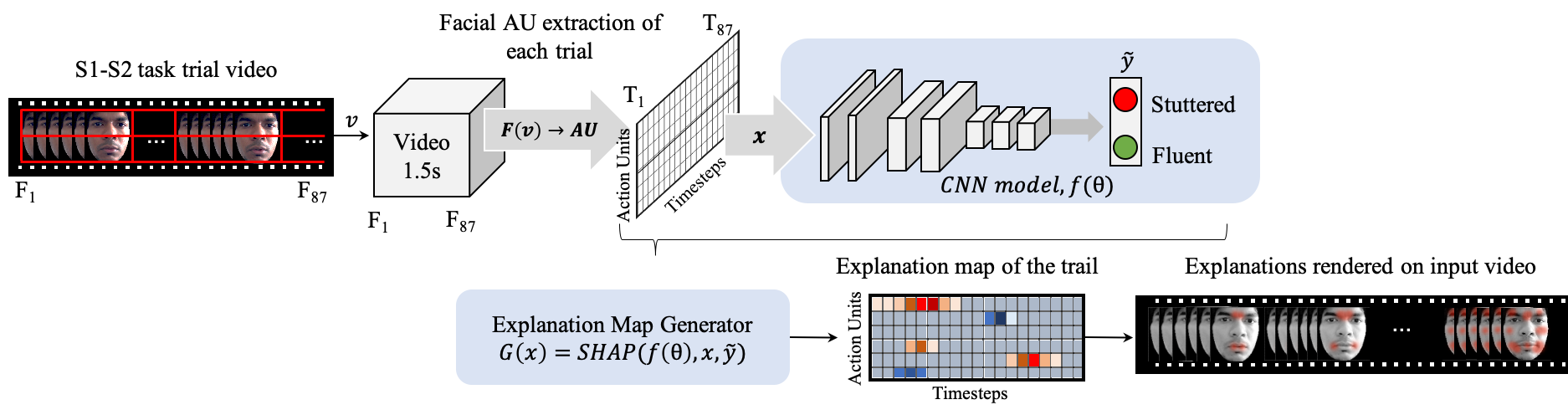}
\caption{The proposed AU-CNN architecture is illustrated with feature dimensions of individual layers. (a) Facial action units are extracted from streaming videos. (b) A CNN model is trained on the extracted AUs. (c) A temporal explanation map is generated for each input sample illustrating AU feature importance for each time window. We show a few time slices of such an explanation map, superimposed on an example face.}
\label{fig:arch}
\end{figure*}

\subsection{CNN Training Scheme: Disfluency Classification using Convolutional Neural Networks}

Primary goals of our classifier, as illustrated in Figure \ref{fig:arch}, is to learn the temporal dynamics of facial muscle movements to predict if a trial was stuttered or fluent and then explain which muscle groups contributed to the classification. Well researched CNN architectures such as ResNet cannot be used in this case due to unequal size of input data dimensions. We design a custom CNN architecture with rectangle kernels and carry out DepthWise and Separable convolutions inspired from EEGNet \cite{Lawhern2018} to learn the temporal changes of AU data. We call this architecture CNN-A. For comparison, we also designed CNN-B similar to the popular VGG-16 architecture with repeated convolution operations with square kernels. We describe our CNN-A architecture here in detail.

Consider $f$ as our CNN-A model to be trained with input data $x \in X_{train}$ with labels $y \in Y_{train}$ where $X_{train}$ and $Y_{train}$ constitute the training dataset. Shape of each $x$ will be $17 \times 87$ as we have 17 AUs and 87 timesteps of information. If $\theta$ is the parameters of the model $f$, our goal for deep learning training is to optimize $\theta$ such that our predictions $\Bar{y} = f(\theta, x)$ are accurate.

We carry out a 2D convolution on the input matrix with $C$ channels and $T$ timesteps. Since we have 17 AUs extracted, $C=17$ for the current study. This is followed by a DepthWise 2D convolution, with a kernel of $(17,1)$ with a depth multiplier of 2 which allows extraction of frequency-specific spatial kernels and reduces the number of parameters in the output. A 2D average pooling reduces the dimensions of the generated feature maps. The output is further passed to a 2D Separable convolution layer, which learns to decouple the relationship within and across feature maps. This is done using a two-step process, where a kernel learns to summarize each feature map individually, and afterwards merging them optimally. 

We feed the output of Separable convolution layer to a Fully Connected Dense layer with 128 neurons. This Dense layer carries the spatial information extracted by prior CNN layers. We call this the embedding layer, which encodes all spatial information. The final classifier layer outputs logits indicating whether a trial is classified as either fluent or stuttered. We use Exponential Linear Unit (ELU) activation function after all convolution and dense layers and use binary cross-entropy as the loss function for the optimization.

CNN-B consists of four convolution layers of filters 16, 32, 64, and 128 respectively, all using (4,4) kernels. Batch normalization and 2D average pooling with a (2,2) filter is applied after each convolution operation. Three Fully Connected Dense layers of size 256, 128, and 64 are used before the final classifier layer. Comparing performance of CNN-B with CNN-A could reveal importance of temporal kernels in CNN-A, if any.

\subsubsection{Actions taken to prevent overfitting issues}
Deep learning networks are non-linear and have many parameters, which can lead to overfitting (i.e. the network learns tiny details and noise in the training set but doesn't generalize to new data). We use a combination of batch normalization, dropout, and early-stopping to curb overfitting issues. Each convolution operation in the proposed network is followed by a batch normalization operation to normalize the activations of the previous layer, effectively reducing the internal covariance shifts that might occur in the architecture. 

Batch normalization technique normalizes the parameters of the deep neural network by multiplying the parameters with the standard deviation and subtracting the mean of parameters. It also works as a regularization for the neural network. Dropout randomly removes the influence of certain neurons towards the output decision, which forces the network to learn better representations of the input. Early stopping is a technique where the training can be stopped according to the training and validation loss. If the validation loss continuously goes higher than training loss, we can stop the training and curb overfitting.

\subsubsection{Training Hyperparameters}
To implement both CNN-A and -B, we used TensorFlow framework \cite{abadi2016tensorflow}. The networks were trained with a batch size of 256 to optimize the parameters using Stochastic Gradient Descend (SGD). Training was scheduled for 500 epochs with early stopping based on validation loss with a patience of 30 epochs. Training started with a high learning rate (LR) of 0.01 and reduced by a factor of 0.5 when the validation loss fails to improve for 15 epochs. This is done using a LR scheduler with a mininum LR of $1e^{-6}$. Logs were generated and saved in Tensorboard.

\subsubsection{Cross-Validation Strategy for Training Data}
In order to make better decisions, the classifier should learn meaningful features and semantics from training data while giving good performance on hold-out testing data which is not used during the training process. To ensure consistent performance, a 5-fold cross-validation strategy was used. Here, after keeping a randomly picked portion of data as the hold-out test set, we picked the training and validation data as five random folds which are equally distributed between stuttered and fluent trials. This 50-50 split on the fluent and stutter trial improves learning decisions and reduces issues related to skewness in data. We measure the performance metrics related to all five training folds to compare the learning performance. All data processing and model training were done on Jetstream research computing cloud \cite{Stewart2015,Towns2014} with NVIDIA GPUs.

\subsubsection{Performance Metrics}
To measure the predictive performance of the proposed deep learning model, we use a combination of accuracy, AUC-ROC which is the Area-under-curve (AUC) of the receiver operating characteristic (ROC), and F1 score. For binary problems, a measure of accuracy is generally not preferred as the main factor of performance due to erroneous interpretations on skewed datasets. Even though we use an equal split between fluent and stutter samples in each dataset for training, validating, and testing our deep learning models, we share other measurements for completeness.

\subsection{Design Considerations for Explainability Algorithm}
\label{xaidesign}
Deep neural networks, despite their impressive performance on real-world datasets are harder to understand and interpret due to their non-linear nature and large number of parameters. Also, due to the black-box nature of deep learning models and other adversarial threats \cite{Chacon2019}, it is often important to understand why a particular decision was made and what fundamental features were used to make that decision \cite{das2020opportunities,Weller2019}. Our CNN models for stuttering prediction using AU data suffer from the same flaws of `black-box' nature of deep neural networks. To explain the muscle groups responsible for a stuttered or fluent speech in an individual trial, we propose the use of DeepSHAP method proposed by Lundberg et al. \cite{Lundberg2017} by utilizing the DeepExplain \cite{Ancona2017} to generate explanations.

Importance of an AU muscle group towards the classifier decision can be understood by studying the feature relevance matrix called attribution (explanation) map. Attribution maps quantify the importance of each AU features towards predicting a fluent or stuttered trial. By adding DeepSHAP as an explainer layer after individual predictions, our final CNN architecture for inference on new samples can be reconsidered as below.

Reconsidering $f$ as our new trained CNN model function and $\theta$ the learnt parameters, with a new input instance $x$, our goal is to predict an accurate solution $\Bar{y}$ and generate an attribution map $g$ which illustrate the feature importance over time with the constraint that shape of $g$ is the same as shape of $x$. Thus, we can generate a correlation between the input $x$ and attribution map $g$. Here, a positive value of attribution for a particular AU at a particular time-window $t$ means that the corresponding AU is highly relevant and is improving the prediction probability of the AI decision around that specific time $t$. Negative values of attribution map $g$ suggests that the feature is decreasing the class probability around a specific time.

Individual trial information once preprocessed generates a matrix with $feature \times timestep$ dimensions specifically,  $17\times87$ for the AU data. The temporal nature of data should be used to generate explanations for specific time-windows to explain the behavior of the model. In order to accomplish this, we define the problem to explain the influence of a set time-window of information towards the particular output. Specifically, for face AU, we encode 17.24 ms of information in each pixel of the input matrix. This is done corresponding to the frequency of data collected under the 1500 ms S1-S2 task. Now, DeepSHAP algorithm can be applied on the input matrix to generate explanation maps which provide feature importance for small time-windows. Hence, each pixel in the explanation map will correspond to the importance of 17.24 ms of the respective AU feature in a specific time of S1-S2 task towards an output prediction.

\section{Experimental Results}
\label{sec:experiments}
In this section we describe the classification results of both CNN-A, CNN-B, and compare them against a random forest (RF) classifier with 500 trees. After comparing the results, we study and statistically analyze, using ANOVA method, the muscle movement attributions of the S1-S2 task across AWS subjects based on their stuttering intensity and also based on different temporal ranges.

\subsection{Classification Results}
\label{subsec:classificationresults}

\begin{figure}[!t]
\centering
\includegraphics[width=\linewidth]{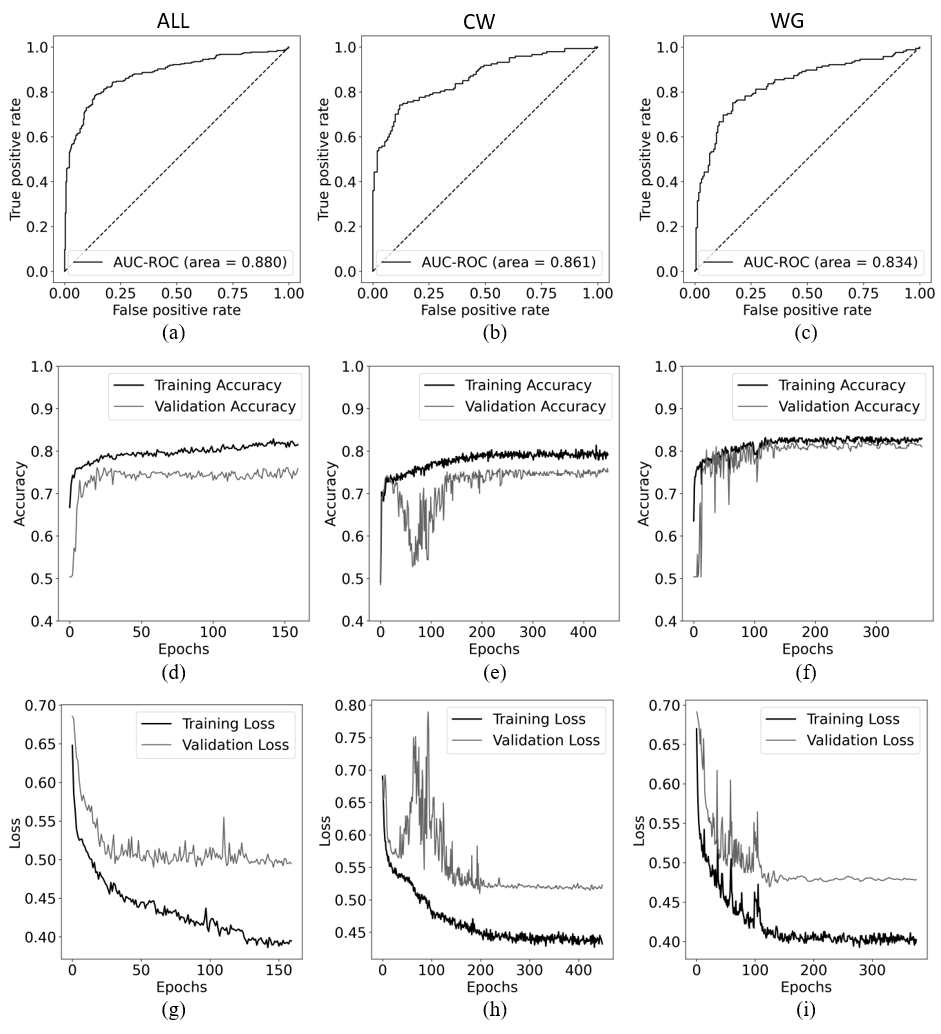}
\caption{Area-under-curve of the receiver operating characteristic (ROC) curve is illustrated for CNN-A for (a) all paradigms (AUC-ROC=0.880), (b) cue-word (AUC-ROC=0.861), and (c) word-go (AUC-ROC=0.834) paradigms respectively; Plot with training and validation accuracies is illustrated for (d) all paradigms, (e) cue-word, and (f) word-go paradigms respectively; Plot with training and validation loss is illustrated for (g) all paradigms, (h) cue-word, and (i) word-go paradigms respectively.}
\label{fig:resultplots}
\end{figure}

Table \ref{tab:results} summarizes how well our CNN-A and CNN-B compares in performance against a random forest (RF) benchmark. Understanding the classifiability of each paradigms (CW and WG) could reveal impact of having the word in memory before speech vocalization and the associated facial muscles. Hence, we train the CNN-A, -B, and RF methods on datasets with all paradigms combined, just CW, and just WG. Figure \ref{fig:resultplots} illustrates AUC-ROC, training and validation accuracy and loss curves of the CNN-A model trained on AU data for the different paradigms. Here, the training accuracy is a measure of how well the model was able to predict on data it had already seen. Validation accuracy is a measure of generalizability and model performance on data which was not included in training. By referring to Figure \ref{fig:resultplots} rows 2 and 3, we can see an optimal fit of each models and infer that the model is not overfitting to the dataset. This can be concluded since training and validation curves follows a downward trend while the corresponding accuracies steadily increases. We can also see that early-stopping algorithm stops the training whenever the validation loss fails to improve. Hence, the number of epochs required for training these models are different for different paradigms.

\begin{table}[!b]
\caption{Performance statistics of CNN-A, -B, and RF baseline trained on a 5-fold cross validation method. Results are presented as mean and standard deviation.}
\label{tab:results}
\centering
\begin{tabular}{|c||cccc|}
\hline
Method & Paradigm      & Accuracy   & AUC-ROC       & F1 Score \\ 
\hline
\hline
\multirow{3}{*}{CNN-A}       & All & \textbf{80.03$\pm$0.55} & \textbf{0.86$\pm$0.01} & \textbf{0.78$\pm$0.01} \\
                             & CW  & 77.30$\pm$1.72 & 0.84$\pm$0.01 & 0.74$\pm$0.02 \\
                             & WG  & 77.88$\pm$0.58 & 0.84$\pm$0.01 & 0.75$\pm$0.01 \\ 
\hline
\multirow{3}{*}{CNN-B}       & All & 76.28$\pm$0.79 & 0.84$\pm$0.00 & 0.75$\pm$0.01 \\
                             & CW  & 70.53$\pm$1.47 & 0.75$\pm$0.02 & 0.67$\pm$0.03 \\
                             & WG  & 76.00$\pm$2.87 & 0.84$\pm$0.03 & 0.76$\pm$0.03 \\ 
\hline
\multirow{3}{*}{RF Baseline} & All & 75.43$\pm$1.16 & 0.75$\pm$0.01 & 0.73$\pm$0.01 \\
                             & CW  & 73.04$\pm$1.66 & 0.73$\pm$0.02 & 0.69$\pm$0.03 \\
                             & WG  & 77.07$\pm$2.83 & 0.77$\pm$0.03 & 0.75$\pm$0.05 \\ 
\hline

\end{tabular}
\end{table}

For CNN-A, when all paradigms are trained together, we see an AUC-ROC value of $0.86\pm0.01$, accuracy of $80.03\pm0.55$\%, and F1-score of $0.78\pm0.01$ suggesting good performance of the classifier on the hold-out test set. CW paradigm alone generated $0.84\pm0.01$ AUC-ROC, an accuracy of $77.30\pm1.72$\%, and F1-score of $0.74\pm0.02$. Considering on the WG paradigm, we see an $0.84\pm0.01$ AUC-ROC, an accuracy of $77.88\pm0.58$\%, and F1-score of $0.75\pm0.01$. Comparing the results with other models, that our proposed CNN-A model outperform CNN-B and random forest baselines by a large margin when all the data is considered. Both CNN-B and RF methods had significant reductions in accuracy and AUC-ROC for CW paradigm.

\begin{table}[!b]
\caption{Impact of changes in kernel size of all convolution operations in CNN-B.}
\label{tab:cnnbdegrade}
\centering
\begin{tabular}{|c||cccc|}
\hline
 Kernel Shape   & Paradigm      & Accuracy   & AUC-ROC       & F1 Score \\ 
\hline
\hline
\multirow{3}{*}{2 $\times$ 2} & All & $\sim$ -2.21\% & $\sim$ -1.99\% & $\sim$ -1.97\% \\
                              & CW  & $\sim$ -2.34\% & $\sim$ -2.10\% & $\sim$ -2.01\% \\
                              & WG  & $\sim$ -0.90\% & $\sim$ -1.01\% & $\sim$ -1.00\% \\ 
\hline
\multirow{3}{*}{6 $\times$ 6} & All & $\sim$ -2.00\% & $\sim$ -1.99\% & $\sim$ -1.99\% \\
                              & CW  & $\sim$ -4.8\% & $\sim$ -1.99\% & $\sim$ -1.98\% \\
                              & WG  & $\sim$ -0.70\% & $\sim$ -1.02\% & $\sim$ -1.02\% \\ 
\hline

\end{tabular}
\end{table}

We argue that data samples from CW paradigm show more microexpressions, which are variations in the muscle movements for short periods of time, rather than macro expressions which are more apparent in WG paradigms. Hence, CW paradigm is harder to classify accurately. CNN-A architecture learns to represent microexpressions better than CNN-B due to the temporal kernels and convolution operations which considers the frequency-specific spatial kernels. To study the impact of different kernel sizes in CNN-B, we carried out an experiment by changing  all kernel sizes to (2,2) and (6,6). This relates to including more temporal information during each convolution operation. However, doing so also includes information from multiple AUs, unlike CNN-A. Table \ref{tab:cnnbdegrade} summarizes the performance degradation caused by changing the kernel sizes. As we can see, no improvement in accuracy was found by lowering or increasing the kernel size.

\subsection{Statistical Significance of Face AUs}
\label{subsec:anova}

Statistical significance study, using methods such as ANOVA, could improve our confidence of trusting the generated explanations. Here, we use an ANOVA model to study the statistical significance of the generated attribution maps to generalize the facial AU patterns of our AWS subjects. We study the significance of holding hypothesis-1, different paradigms in final classification, stuttering rate especially high-stutter-rate (HSR, stutter rate$>$40\%) and low-stutter-rate (LSR, stutter rate$<$40\%) subjects during speech preparation, and the significance of temporal information in specific time-windows. To understand the impact of specific time-windows in fluent and stuttered trials, we split the temporal range into a) 0-500 ms, b) 500-800 ms, and c) 1100-1500 ms.

\begin{figure*}[!t]
\centering
\includegraphics[width=0.9\linewidth]{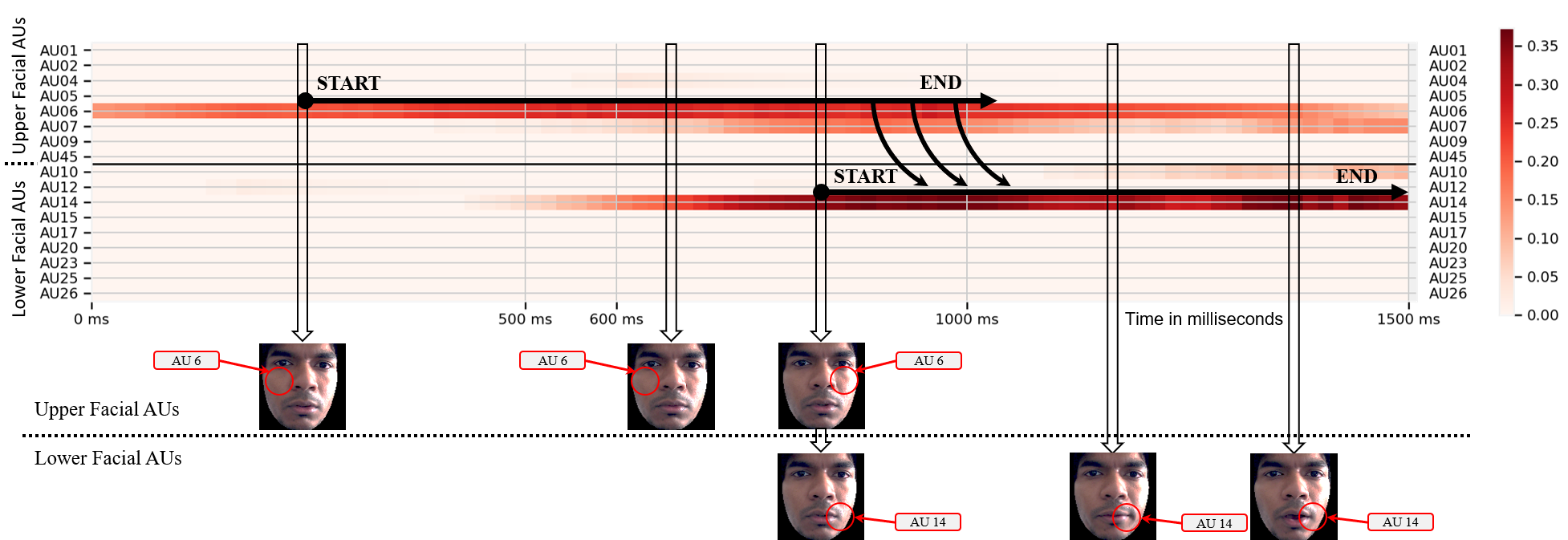}
\caption{Explanation map for a single stutter trial from a test sample of AU data collected from the first author attempting the S1-S2 task. For visualization purposes, we filter out the negative attributions. The black line in the middle splits the Y axis to upper facial region (until AU45) and lower facial region (from AU 12 to end), respectively. After removing all features with negative correlations, We can see a high feature importance (red) for AU 6 (cheek raiser), AU 7 (eyelid tightener), AU 10 (upper lip raiser), AU 14 (dimpler).}
\label{fig:explanationmap}
\end{figure*}

\subsection{Explaining the Classifier Decisions}
\label{subsec:explainations}

Figure \ref{fig:explanationmap} illustrates an explanation map (showing only the positive attributions to aid visualization) generated for a single stutter trial from a test sample of AU data collected from the first author attempting the S1-S2 task. Here, individual AUs are populated in the y-axis and timesteps in the x-axis. Upper and lower AUs are separated in the middle by a black line. Each square pixel in the explanation map corresponds to explanations generated for 17.24 ms of AU data as described in Section \ref{xaidesign}. Explanation maps, similar to Figure \ref{fig:explanationmap} with both positive and negative attributions, are generated for all trials in the hold-out test set and a database of explanation maps were generated. Due to space constraints, further discussions will use ANOVA results or line plots of AUs extracted from similar explanation maps to aid the discussion. 

\subsubsection*{Significance of holding hypothesis 1}
Statistical analysis on the facial regions of all subjects revealed high significance of AU 6 (cheek raiser, F=23.47, p$<$0.005) and AU 4 (brow lowerer, F=19.17, p$<$0.005) from upper facial region, and AU 14 (dimpler, F=80.3, p$<$0.005), AU 10 (upper lip raiser, F=53.33, p$<$0.005), and AU 12 (lip corner puller F=27.19, p$<$0.005) from lower facial region. We evaluated our hypothesis that pre-speech facial activity can encode enough information to be classified as fluent or stuttered trials using the CNN models, of which the results are described in Table \ref{tab:results}. Furthermore, we find that the average attributions of all subjects for stutter trials, as illustrated in Figure \ref{fig:resultfig} (a), for AU 6 and 14 which are significant across all subjects, are consistently higher than that of the fluent trials. This further strengthens our evaluation of the hypothesis using attribution maps generated using explainable AI methods.

\subsubsection*{Significance of paradigms}
Statistical analysis on the attribution values indicated a weak statistical importance of paradigms (F=4.8, p$<$0.05) CW and WG in determining the trial as fluent or stuttered. However, this can neither be generalized nor neglected due to the weak significance. Further study needs to be done on a larger population of subjects to find a conclusion. Hence, our preliminary study shows that having the words in memory prior to speech preparation does not significantly impact the facial muscle responses but shows weak statistical correlation during speech preparation. Even-though we find microexpressions in CW and larger active muscle movements in WG, more data is required to generalize the results. Hence, a part of hypothesis 4 regarding the impact of speech plan in short-term memory does not hold completely.

\subsubsection*{Significance of stutter rate}
Stutter-rate of individual subjects revealed important findings and proved to be statistically significant in predicting stuttering disfluency (F=18.66, p$<$0.005). HSR subjects showed statistical significance for more AUs on both upper (AU 4: brow lowerer, 6: cheek raiser, 7: eyelid tightener, 45: eye blinker) and lower (AU 14: dimpler, 15: lip corner depressor, 17: chin raiser, 20: lip stretcher) facial regions compared to lower stutter rate (LSR). These results suggest that there are larger number upper and lower muscle groups contributing towards stuttered than fluent trials. Relatedly, we could infer that subjects who stuttered more on the speech preparation task had comparatively higher facial muscle activity than subjects who stuttered less.

\subsubsection*{Significance of time-windows}
\begin{figure*}[!t]
\centering
\includegraphics[width=0.9\linewidth]{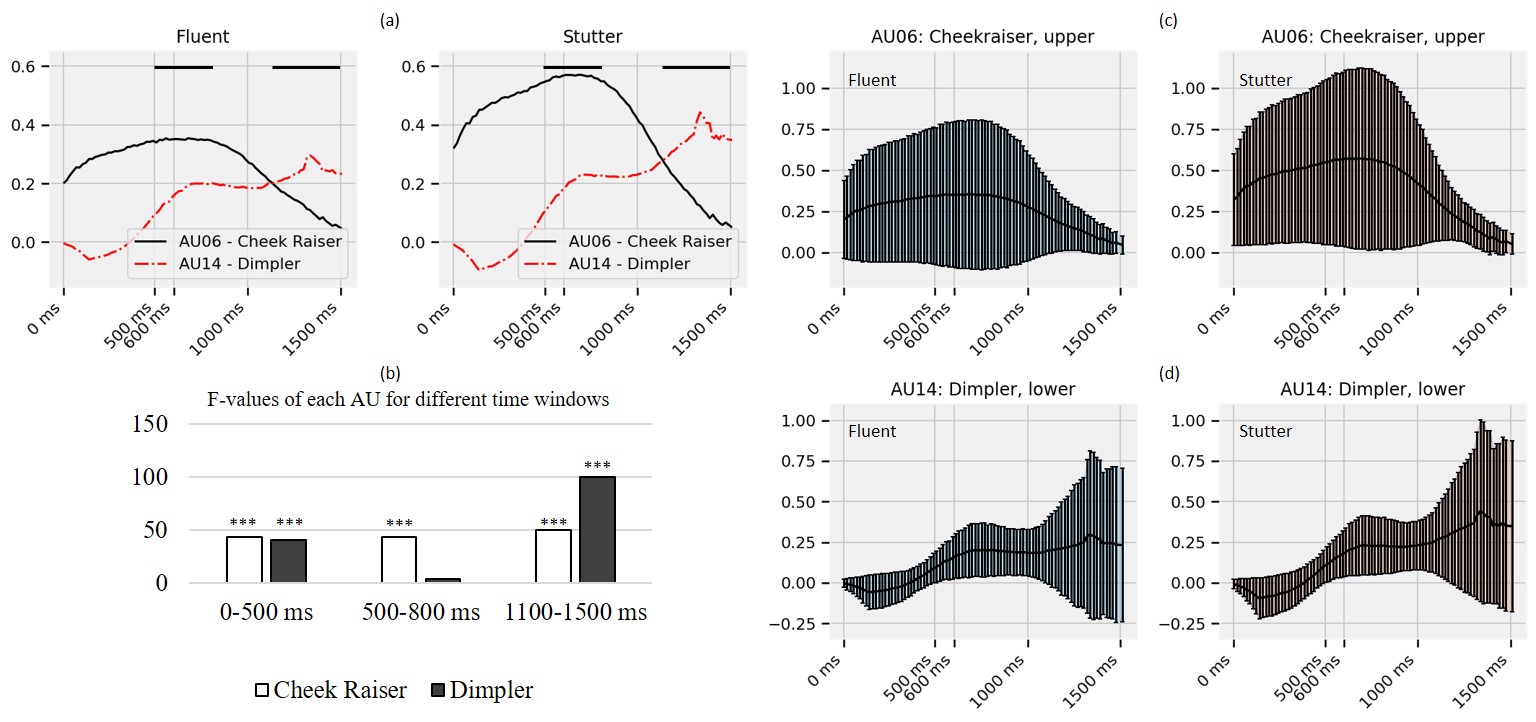}
\caption{Average attributions for fluent and stutter trials for AU 6 (cheek raiser) and AU 14 (dimpler) are illustrated along with statistical significance of specific time-windows. Figure (a) compares the average attributions side by side. Figure (b) illustrate the statistical significance of time-windows. Figures (c) and (d) illustrates the standard deviation of results for fluent and stutter trials.}
\label{fig:resultfig}
\end{figure*}

Our analysis suggests that AU 6 and AU 14 are consistently significant considering all subjects. Hence, we focus on AUs 6 and 14 for this study. As illustrated in Figure \ref{fig:resultfig} (b), we see that AU 6 is statistically significant across all time-windows under consideration (a: F=43.91, p$<$0.001, b: F=43.86, p$<$0.001, c: F=50.1, p$<$0.001). Also, AU 14 is significant across 0-500 ms (F=40.83, p$<$0.001) and 1100-1500 ms (F=100.49, p$<$0.001) while statistically insignificant in 500-800ms (F=4.37, p$>$0.05) considering all subjects and trials. This correlates with the upward trend of AU 6 which peaks around 700 ms and drops rapidly towards 1500 ms showing high importance of upper facial muscles towards the point when the cue is removed at 1000 ms. Also, magnitude of attributions are higher in stutter trials than fluent trials as illustrated in Figure \ref{fig:resultfig} (c). Hence, we can confirm hypothesis 3 that upper facial muscles show greater activity in stutter vs. fluent trials. 

In contrast, attributions of AU 14 have a negative trend at first and starts to improve towards 500 ms. During 500-1000 ms, there is very little change in the impact AU 14 has towards classifying stutter and fluent trials as illustrated in Figure \ref{fig:resultfig} (d). However, during the 1100-1500 ms, just before the end of the trial, we see a large jump in the lower facial activity with considerable statistical significance. Also, we see larger magnitude of attributions for stutter trials vs. fluent trials indicating that parts of hypothesis 4 regarding peak activity towards S2 and greater activity for stutter trials holds true.  This tends to show a behavioral pattern where importance of upper facial muscles is high at S1 and reduces towards S2, which could indicate more focused attention at S1 (hypothesis 3), while lower facial muscles improves towards S2 after being dormant during S1, which could indicate a motor preparation to speak after S2 (hypothesis 4) as illustrated in Figure \ref{fig:explanationmap}.

\section{Discussions}
\label{sec:discussions}
This work demonstrates an experimental framework, machine learning methods, and explainable AI algorithms capable of predicting solely from facial muscle movements whether an AWS will be fluent or stutter before they speak. The video data emphasized the usability of pre-speech facial activity while also confirming our primary hypothesis that pre-speech facial activity contains enough information to accurately differentiate in advance whether speech will be fluent or stuttered. This is a major finding that has many clinical implications. For instance, this model can be used to identify stuttering moments in real-time along with the presence of any observable secondary behaviors related to facial tension to estimate not only the frequency of stuttering, but also to estimate severity for assessment purposes. The same method can further be used in conjunction with behavioral therapy to detect stuttering moments and any associated facial tension to provide non-invasive feedback using standard hardware available on most digital devices (computers, tablets, smartphones). 

Behavioral stuttering therapy incorporates either fluency shaping or stuttering modification/management strategies, or a combination of both to help AWS increase fluency and reduce stuttering severity \cite{prins2009evidence}. Use of real-time feedback based on facial AUs can help with the identification stage of stuttering modification and also aid in providing real-time feedback to help reduce facial tension to decrease stuttering severity and increase overall fluency, similar to the effect seen from using facial EMG \cite{Hancock1998}. Clinical applications would need to be tested exhaustively in controlled experiments; however, the possibility is promising considering the evidence supporting the efficacy of EMG based bio-feedback \cite{Hancock1998} that required specialized equipment that is invasive. 

The speech patterns of both fluent and stuttered speech along with CW and WG task variations showed very similar attribution patterns. Nonetheless, the data still supported our second hypothesis that disfluent speech trials would show higher facial AUs. The stuttered trials showed large variance and higher average response when compared to fluent trials. Attributions of two AUs (AUs 6 \& 14) that peaked at different times appear to show the most promise for differentiating between future fluent or stuttered speech. Facial AU 6 (cheek raiser) from upper face increased rapidly from S1 up until 800 ms after S1, with a significantly higher peak in the stuttered vs. fluent speech trials. We suggest this upper face AU is related to the arousal associated with the start of each trial due to activity being driven by S1 onset and no observed difference between the WG and CW paradigms. High arousal is known to influence stuttering rates. For instance, public speaking can induce increased disfluencies in both people who stutter and fluent individuals \cite{bodie2010racing}.

Facial AU 14 (dimpler) from lower face supported our hypothesis by showing increased attribution and variability before stuttered vs. fluent speech. Facial AU 14 starts to slowly increase after S1 until a rapid increase occurs right before S2, independent of paradigm. These finding also partially supports our hypotheses that lower facial AUs around the lips can index the anticipation of speaking by showing a rapid increase right before S2 onset which prompts the subject to speak. However, since there was no significant difference between the CW and WG tasks, the lower facial AUs don’t appear to index whether a specific speech plan is being held in short-term memory.

The combination of these two AU, one for lower and upper facial regions also supports our hypothesis that facial AUs may be able to independently separate the amount of emotional arousal (upper AUs) and anticipation to speak (lower facial AUs).  Our previous EEG research suggests brain states conducive to fluent or stuttered speech can change in short time frames (i.e. seconds) \cite{Myers2019}. The non-word pairs were designed not to induce an emotional arousal, yet attributions of upper facial movements appear to capture such rapid shifts in arousal irrespective of S1 being a non-word pair or a  cross that lets the subject know when S2 will appear. The ability to distinguish fluent for disfluent trials during speech preparation irrespective of paradigm (WG or CW) provides further evidence that moments of stuttering are triggered not just by what is about to be said but instead the brain state the person who stutters is trying to speak through.

\section{Conclusions and Future Work}
\label{sec:conclusion}
This paper presents an XAI-assisted novel speech disfluency study to investigate the usability of facial muscle movements before speech vocalization to automatically classify upcoming speech as fluent or stuttered. Our study revealed temporal facial muscle movement patterns in the upper and lower facial regions of the face which indicate anticipation and speech preparation during our S1-S2 task which are related to stuttering. Consistently high correlation of AU6 (cheek raiser) and AU 14 (dimpler) along with similar but highly variable attribution values suggests that AWS have larger facial activity before speech vocalization of stutter trials, which the AI algorithms can identify.

The results of this novel study proved our hypothesis that pre-speech facial activities encode enough information to distinguish if a future speech vocalization would be fluent or stuttered. However, a comprehensive study on larger cohort of AWS is required to generalize the idea and assess the broader impact of having the word to be spoken in short-term memory. With enough data, future studies could also find personalized facial muscle patterns or traits in each subject which we have aggregated during the data processing of the current study. Multi-modal studies including EEG and eye-tracker information could study correlations between brain activity and facial muscle movements of AWS to reveal cognitive and behavioral markers leading to stuttering.


\ifCLASSOPTIONcaptionsoff
  \newpage
\fi


\bibliographystyle{IEEEtran}
\bibliography{references}

\begin{thebibliography}{10}
\providecommand{\url}[1]{#1}
\csname url@samestyle\endcsname
\providecommand{\newblock}{\relax}
\providecommand{\bibinfo}[2]{#2}
\providecommand{\BIBentrySTDinterwordspacing}{\spaceskip=0pt\relax}
\providecommand{\BIBentryALTinterwordstretchfactor}{4}
\providecommand{\BIBentryALTinterwordspacing}{\spaceskip=\fontdimen2\font plus
\BIBentryALTinterwordstretchfactor\fontdimen3\font minus
  \fontdimen4\font\relax}
\providecommand{\BIBforeignlanguage}[2]{{%
\expandafter\ifx\csname l@#1\endcsname\relax
\typeout{** WARNING: IEEEtran.bst: No hyphenation pattern has been}%
\typeout{** loaded for the language `#1'. Using the pattern for}%
\typeout{** the default language instead.}%
\else
\language=\csname l@#1\endcsname
\fi
#2}}
\providecommand{\BIBdecl}{\relax}
\BIBdecl

\bibitem{manning2017clinical}
W.~H. Manning and A.~DiLollo, \emph{Clinical decision making in fluency
  disorders}.\hskip 1em plus 0.5em minus 0.4em\relax Plural Publishing, 2017.

\bibitem{prasse2008stuttering}
J.~E. Prasse and G.~E. Kikano, ``Stuttering: an overview,'' \emph{American
  family physician}, vol.~77, no.~9, pp. 1271--1276, 2008.

\bibitem{Myers2019}
J.~Myers, F.~Irani, E.~Golob, J.~Mock, and K.~Robbins, ``{Single-Trial
  Classification of Disfluent Brain States in Adults Who Stutter},''
  \emph{Proc. - 2018 IEEE Int. Conf. Syst. Man, Cybern. SMC 2018}, pp. 57--62,
  2019.

\bibitem{ekman1997face}
P.~Ekman and E.~L. Rosenberg, \emph{{What the Face RevealsBasic and Applied
  Studies of Spontaneous Expression Using the Facial Action Coding System
  (FACS)}}.\hskip 1em plus 0.5em minus 0.4em\relax Oxford University Press, apr
  2005.

\bibitem{friesen1978facial}
E.~Friesen and P.~Ekman, ``{Facial action coding system: a technique for the
  measurement of facial movement},'' \emph{Palo Alto}, vol.~3, 1978.

\bibitem{Sayette2002}
M.~A. Sayette, J.~F. Cohn, J.~M. Wertz, M.~A. Perrott, and D.~J. Parrot, ``{A
  Psychometric Evaluation of the Facial Action Coding System for Assessing
  Spontaneous Expression Michael A. Sayette, Jeffrey F. Cohn, Joan M. Wertz,
  Michael A. Perrott, and Dominic J. Parrott University of Pittsburgh},''
  \emph{J. Nonverbal Behav.}, 2002.

\bibitem{Hamm2011}
J.~Hamm, C.~G. Kohler, R.~C. Gur, and R.~Verma, ``{Automated Facial Action
  Coding System for dynamic analysis of facial expressions in neuropsychiatric
  disorders},'' \emph{J. Neurosci. Methods}, vol. 200, no.~2, pp. 237--256, sep
  2011.

\bibitem{Lints-Martindale2007}
A.~C. Lints-Martindale, T.~Hadjistavropoulos, B.~Barber, and S.~J. Gibson, ``{A
  Psychophysical Investigation of the Facial Action Coding System as an Index
  of Pain Variability among Older Adults with and without Alzheimer's
  Disease},'' \emph{Pain Med.}, vol.~8, no.~8, pp. 678--689, nov 2007.

\bibitem{Wang2013}
Z.~Wang, S.~Wang, and Q.~Ji, ``{Capturing Complex Spatio-temporal Relations
  among Facial Muscles for Facial Expression Recognition},'' in \emph{2013 IEEE
  Conf. Comput. Vis. Pattern Recognit.}\hskip 1em plus 0.5em minus 0.4em\relax
  IEEE, jun 2013, pp. 3422--3429.

\bibitem{Wehrle2000}
T.~Wehrle, S.~Kaiser, S.~Schmidt, and K.~R. Scherer, ``{Studying the dynamics
  of emotional expression using synthesized facial muscle movements.}''
  \emph{J. Pers. Soc. Psychol.}, vol.~78, no.~1, pp. 105--119, 2000.

\bibitem{Mehu2012}
M.~Mehu, M.~Mortillaro, T.~B{\"{a}}nziger, and K.~R. Scherer, ``{Reliable
  facial muscle activation enhances recognizability and credibility of
  emotional expression.}'' \emph{Emotion}, vol.~12, no.~4, pp. 701--715, aug
  2012.

\bibitem{meng2017listen}
Z.~Meng, S.~Han, and Y.~Tong, ``Listen to your face: Inferring facial action
  units from audio channel,'' \emph{IEEE Transactions on Affective Computing},
  2017.

\bibitem{Zhang2013}
J.~Zhang, B.~Dong, and Y.~Yan, ``{A computer-assist algorithm to detect
  repetitive stuttering automatically},'' \emph{Proc. - 2013 Int. Conf. Asian
  Lang. Process. IALP 2013}, pp. 249--252, 2013.

\bibitem{Villegas2019}
B.~Villegas, K.~M. Flores, K.~{Jose Acuna}, K.~Pacheco-Barrios, and D.~Elias,
  ``{A Novel Stuttering Disfluency Classification System Based on Respiratory
  Biosignals},'' in \emph{2019 41st Annu. Int. Conf. IEEE Eng. Med. Biol.
  Soc.}\hskip 1em plus 0.5em minus 0.4em\relax IEEE, jul 2019, pp. 4660--4663.

\bibitem{Teixeira2017}
J.~P. Teixeira, M.~G. Fernandes, and R.~A. Costa, ``{Automatic Determination of
  Pauses in Speech for Classification of Stuttering Disorder},'' in
  \emph{Design, Development, and Integration of Reliable Electronic Healthcare
  Platforms}, 2017, pp. 132--149.

\bibitem{Mahesha2017}
P.~Mahesha and D.~S. Vinod, ``{LP-Hillbert transform based MFCC for effective
  discrimination of stuttering dysfluencies},'' in \emph{2017 Int. Conf. Wirel.
  Commun. Signal Process. Netw.}\hskip 1em plus 0.5em minus 0.4em\relax IEEE,
  mar 2017, pp. 2561--2565.

\bibitem{das2020opportunities}
A.~Das and P.~Rad, ``Opportunities and challenges in explainable artificial
  intelligence (xai): A survey,'' \emph{arXiv preprint arXiv:2006.11371}, 2020.

\bibitem{holzinger2017we}
A.~Holzinger, C.~Biemann, C.~S. Pattichis, and D.~B. Kell, ``What do we need to
  build explainable ai systems for the medical domain?'' \emph{arXiv preprint
  arXiv:1712.09923}, 2017.

\bibitem{lundberg2017unified}
S.~M. Lundberg and S.-I. Lee, ``A unified approach to interpreting model
  predictions,'' in \emph{Advances in neural information processing systems},
  2017, pp. 4765--4774.

\bibitem{tan20193d}
J.~Tan, Y.~Gao, Z.~Liang, W.~Cao, M.~J. Pomeroy, Y.~Huo, L.~Li, M.~A. Barish,
  A.~F. Abbasi, and P.~J. Pickhardt, ``3d-glcm cnn: A 3-dimensional gray-level
  co-occurrence matrix based cnn model for polyp classification via ct
  colonography,'' \emph{IEEE Transactions on Medical Imaging}, 2019.

\bibitem{mock2015speech}
J.~R. Mock, A.~L. Foundas, and E.~J. Golob, ``Speech preparation in adults with
  persistent developmental stuttering,'' \emph{Brain and language}, vol. 149,
  pp. 97--105, 2015.

\bibitem{jeffmock2016cortical}
A.~L. Mock, Jeffrey R~Foundas and E.~J. Golob, ``Cortical activity during cued
  picture naming predicts individual differences in stuttering frequency,''
  \emph{Clinical Neurophysiology}, vol. 127, no.~9, pp. 3093--3101, 2016.

\bibitem{sheehan1974stuttering}
J.~G. Sheehan, ``Stuttering behavior: A phonetic analysis,'' \emph{Journal of
  Communication Disorders}, vol.~7, no.~3, pp. 193--212, 1974.

\bibitem{bloodstein2008handbook}
O.~Bloodstein and N.~B. Ratner, \emph{A handbook on stuttering.}\hskip 1em plus
  0.5em minus 0.4em\relax Clifton Park (N.Y.): Thomson/Delmar Learning, 2008.

\bibitem{Baltrusaitis2015}
T.~Baltrusaitis, M.~Mahmoud, and P.~Robinson, ``{Cross-dataset learning and
  person-specific normalisation for automatic Action Unit detection},'' in
  \emph{2015 11th IEEE Int. Conf. Work. Autom. Face Gesture Recognit.}\hskip
  1em plus 0.5em minus 0.4em\relax IEEE, may 2015, pp. 1--6.

\bibitem{samad2017feasibility}
M.~D. Samad, N.~Diawara, J.~L. Bobzien, J.~W. Harrington, M.~A. Witherow, and
  K.~M. Iftekharuddin, ``A feasibility study of autism behavioral markers in
  spontaneous facial, visual, and hand movement response data,'' \emph{IEEE
  Transactions on Neural Systems and Rehabilitation Engineering}, vol.~26,
  no.~2, pp. 353--361, 2017.

\bibitem{Lawhern2018}
V.~J. Lawhern, A.~J. Solon, N.~R. Waytowich, S.~M. Gordon, C.~P. Hung, and
  B.~J. Lance, ``{EEGNet: a compact convolutional neural network for EEG-based
  brain–computer interfaces},'' \emph{J. Neural Eng.}, vol.~15, no.~5, p.
  056013, oct 2018.

\bibitem{abadi2016tensorflow}
M.~Abadi, P.~Barham, J.~Chen, Z.~Chen, A.~Davis, J.~Dean, M.~Devin,
  S.~Ghemawat, G.~Irving, M.~Isard \emph{et~al.}, ``Tensorflow: A system for
  large-scale machine learning,'' in \emph{12th $\{$USENIX$\}$ symposium on
  operating systems design and implementation ($\{$OSDI$\}$ 16)}, 2016, pp.
  265--283.

\bibitem{Stewart2015}
C.~A. Stewart, G.~Turner, M.~Vaughn, N.~I. Gaffney, T.~M. Cockerill, I.~Foster,
  D.~Hancock, N.~Merchant, E.~Skidmore, D.~Stanzione, J.~Taylor, and S.~Tuecke,
  ``{Jetstream},'' in \emph{Proc. 2015 XSEDE Conf. Sci. Adv. Enabled by Enhanc.
  Cyberinfrastructure - XSEDE '15}.\hskip 1em plus 0.5em minus 0.4em\relax New
  York, New York, USA: ACM Press, 2015, pp. 1--8.

\bibitem{Towns2014}
J.~Towns, T.~Cockerill, M.~Dahan, I.~Foster, K.~Gaither, A.~Grimshaw,
  V.~Hazlewood, S.~Lathrop, D.~Lifka, G.~D. Peterson, R.~Roskies, J.~R. Scott,
  and N.~Wilkins-Diehr, ``{XSEDE: Accelerating Scientific Discovery},''
  \emph{Comput. Sci. Eng.}, vol.~16, no.~5, pp. 62--74, sep 2014.

\bibitem{Chacon2019}
H.~Chacon, S.~Silva, and P.~Rad, ``{Deep Learning Poison Data Attack
  Detection},'' in \emph{2019 IEEE 31st Int. Conf. Tools with Artif.
  Intell.}\hskip 1em plus 0.5em minus 0.4em\relax IEEE, nov 2019, pp. 971--978.

\bibitem{Weller2019}
A.~Weller, ``{Transparency: Motivations and Challenges},'' in \emph{Lect. Notes
  Comput. Sci. (including Subser. Lect. Notes Artif. Intell. Lect. Notes
  Bioinformatics)}.\hskip 1em plus 0.5em minus 0.4em\relax Springer, Cham,
  2019, vol. 11700 LNCS, pp. 23--40.

\bibitem{Lundberg2017}
S.~M. Lundberg and S.~I. Lee, ``{A unified approach to interpreting model
  predictions},'' in \emph{Adv. Neural Inf. Process. Syst.}, 2017.

\bibitem{Ancona2017}
M.~Ancona, E.~Ceolini, C.~{\"{O}}ztireli, and M.~Gross, ``{Towards better
  understanding of gradient-based attribution methods for Deep Neural
  Networks},'' \emph{6th Int. Conf. Learn. Represent. ICLR 2018 - Conf. Track
  Proc.}, nov 2017.

\bibitem{prins2009evidence}
D.~Prins and R.~J. Ingham, ``Evidence-based treatment and
  stuttering—historical perspective,'' \emph{Journal of Speech, Language, and
  Hearing Research}, 2009.

\bibitem{Hancock1998}
K.~Hancock, A.~Craig, C.~McCready, A.~McCaul, D.~Costello, K.~Campbell, and
  G.~Gilmore, ``{Two- to Six-Year Controlled-Trial Stuttering Outcomes for
  Children and Adolescents},'' \emph{J. Speech, Lang. Hear. Res.}, vol.~41,
  no.~6, pp. 1242--1252, dec 1998.

\bibitem{bodie2010racing}
G.~D. Bodie, ``A racing heart, rattling knees, and ruminative thoughts:
  Defining, explaining, and treating public speaking anxiety,''
  \emph{Communication education}, vol.~59, no.~1, pp. 70--105, 2010.

\end{thebibliography}



\end{document}